\documentclass[runningheads]{llncs}
\usepackage[pdftex]{graphicx}
\usepackage{xcolor}
\usepackage{multirow}
\usepackage{epstopdf}
\usepackage{xcolor}
\usepackage{multirow}
\usepackage{hyperref}
\usepackage[caption=false,font=normalsize,labelfont=sf,textfont=sf, position=top]{subfig}

\begin{document}
\pagestyle{headings}
\mainmatter


\title{Semantic Segmentation with Reverse Attention} 

\author{Qin Huang, Chunyang Xia, Wuchi Hao, Siyang Li, Ye Wang, Yuhang Song and C.-C. Jay Kuo}
\institute{(qinhuang@usc.edu)}

\maketitle


\begin{abstract}
Recent development in fully convolutional neural network enables efficient end-to-end learning of
semantic segmentation. Traditionally,  the convolutional classifiers are taught to learn the representative semantic features
of labeled semantic objects.  In this work, we
propose a reverse attention network (RAN) architecture that trains the
network to capture the opposite concept (i.e., what are not associated
with a target class) as well. The RAN is a three-branch network that
performs the direct, reverse and reverse-attention learning processes
simultaneously.  Extensive experiments are conducted to show the
effectiveness of the RAN in semantic segmentation.  Being built upon the
DeepLabv2-LargeFOV, the RAN achieves the state-of-the-art mean IoU score
($48.1\%$) for the challenging PASCAL-Context dataset.  Significant
performance improvements are also observed for the PASCAL-VOC, Person-Part, NYUDv2 and ADE20K datasets. 
\end{abstract}

\section{Introduction}\label{sec:intro}

Semantic segmentation is an important task for image understanding and
object localization.  With the development of fully-convolutional neural
network (FCN) \cite{long2015fully}, there has been a significant
advancement in the field using end-to-end trainable networks.  The
progress in deep convolutional neural networks (CNNs) such as the VGGNet
\cite{simonyan2014very}, Inception Net \cite{szegedy2015going}, and
Residual Net \cite{he2015deep} pushes the semantic segmentation
performance even higher via comprehensive learning of high-level
semantic features.  Besides deeper networks, other ideas have been
proposed to enhance the semantic segmentation performance. For example,
low-level features can be explored along with the high-level semantic
features \cite{bishop2006pattern} for performance
improvement.  Holistic image understanding can also be used to boost the
performance \cite{lin2016refinenet,zhao2016pyramid,hu2016recalling}.
Furthermore, one can guide the network learning by generating
highlighted targets \cite{doersch2015unsupervised, dai2016instance, dai2016r,shrivastava2016training,wu2016high,wu2016bridging}. Generally
speaking, a CNN can learn the semantic segmentation task more
effectively under specific guidance. 

\begin{figure*}
\begin{center}
\includegraphics[width=12.8cm]{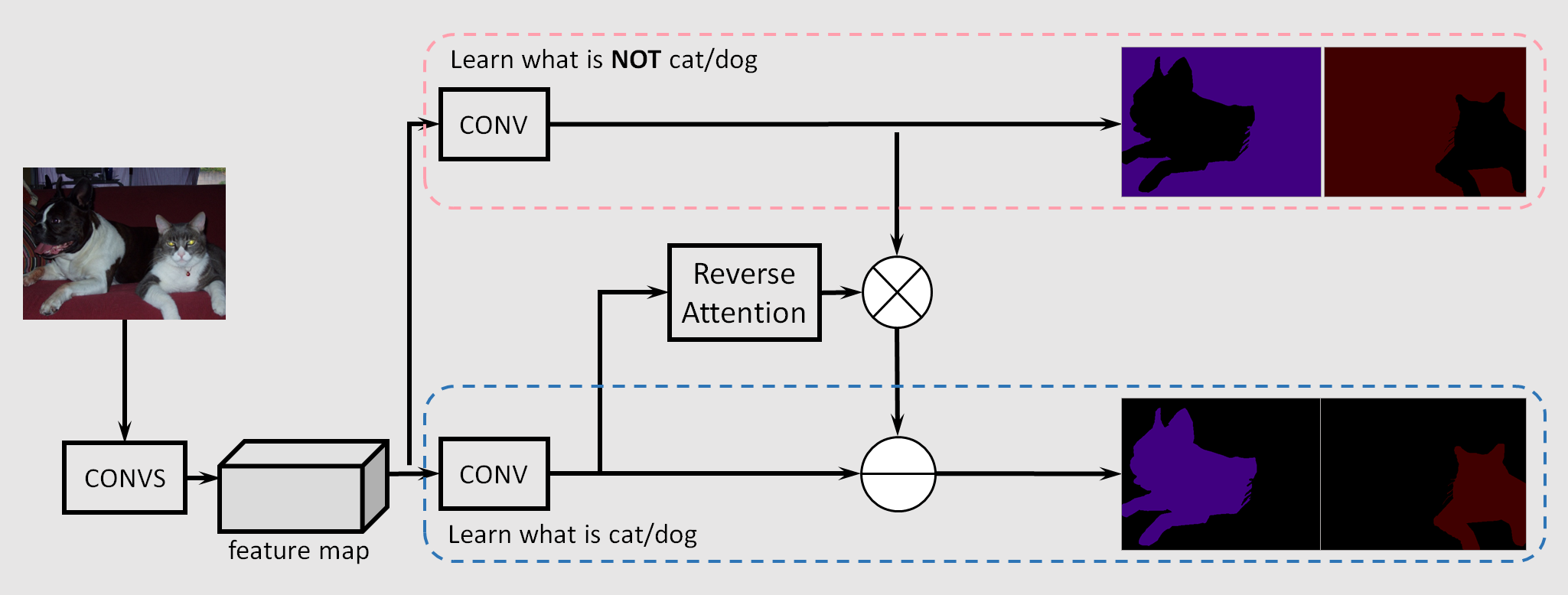}
\end{center}
\caption{An illustration of the proposed reversed attention network (RAN),
where the lower and upper branches learn features and predictions that
are and are not associated with a target class, respectively.  The
mid-branch focuses on local regions with complicated spatial patterns
whose object responses are weaker and provide a mechanism to amplify
the response. The predictions of all three branches are fused to yield
the final prediction for the segmentation task.} \label{fig:overview}
\end{figure*}



In spite of these developments, all existing methods focus on the
understanding of the features and prediction of the target class.
However, there is no mechanism to specifically teach the network to learn the
difference between classes. The high-level semantic features are
sometimes shared across different classes (or between an object and its
background) due to a certain level of visual similarity among classes in
the training set.  This will yield a confusing results in regions that
are located in the boundary of two objects (or object/background) since
the responses to both objects (or an object and its background) are
equally strong. Another problem is caused by the weaker responses of the
target object due to a complicated mixture of objects/background. It is
desirable to develop a mechanism to identify these regions and amplify
the weaker responses to capture the target object. We are not aware of
any effective solution to address these two problems up to now. In this
work, we propose a new semantic segmentation architecture called the reverse attention
network (RAN) to achieve these two goals. A conceptual overview of the
RAN system is shown in Fig. \ref{fig:overview}. 

The RAN uses two separate branches to learn features and generate predictions
that are and are not associated with a target class, respectively. To further highlight the knowledge learnt from reverse-class, we design a reverse attention structure, which generates per-class mask to amplify the reverse-class response in the confused region. The predictions of all three branches are finally
fused together to yield the final prediction for the segmentation task.
We build the RAN upon the state-of-the-art Deeplabv2-LargeFOV with the
ResNet-101 structure and conduct comprehensive experiments on many datasets, including
PASCAL VOC, PASCAL Person Part, PASCAL Context, NYU-Depth2, and ADE20K
MIT datasets.  Consistent and significant improvements across the
datasets are observed. We implement the proposed RAN in Caffe \cite{jia2014caffe}, and the trained network structure with models are
available to the public 
\footnote{\url{https://drive.google.com/drive/folders/0By2w_A-aM8Rzbllnc3JCQjhHYnM?usp=sharing}}. 

\section{Related Work}\label{review}
 
A brief review on recent progresses in semantic segmentation is given in
this section. Semantic segmentation is a combination of the pixel-wisea
localization task \cite{zhang1996survey,shi2000normalized} and the
high-level recognition task.  Recent developments in deep CNNs
\cite{krizhevsky2012imagenet,simonyan2014very,szegedy2015going} enable
comprehensive learning of semantic features using a large amount of
image data \cite{Everingham10, lin2014microsoft,deng2009imagenet}.  The
FCN \cite{long2015fully} allows effective end-to-end learning by
converting fully-connected layers into convolutional layers.  

Performance improvements have been achieved by introducing several new
ideas.  One is to integrate low- and high-level convolutional features
in the network. This is motivated by the observation that the pooling
and the stride operations can offer a larger filed of view (FOV) and
extract semantic features with fewer convolutional layers, yet it
decreases the resolution of the response maps and thus suffers from
inaccurate localization. The combination of segmentation results from
multiple layers was proposed in \cite{long2015fully,
shuai2016improving}.  Fusion of multi-level features before decision
gives an even better performance as shown in \cite{chen2016attention,
lin2016refinenet}.  Another idea, as presented in
\cite{chen2016deeplab}, is to adopt a dilation
architecture to increase the resolution of response maps while
preserving large FOVs. In addition, both local- and long-range
conditional random fields can be used to refine segmentation details as
done in \cite{zheng2015conditional, chen2016semantic}.
Recent advances in the RefineNet \cite{lin2016refinenet} and the PSPNet
\cite{zhao2016pyramid} show that a holistic understanding of the whole
image \cite{hu2016recalling} can boost the segmentation performance
furthermore. 

Another class of methods focuses on guiding the learning procedure with
highlighted knowledge. For example, a hard-case learning was adopted in
\cite{shrivastava2016training} to guide a network to focus on less
confident cases.  Besides, the spatial information can be explored to
enhance features by considering coherence with neighboring patterns
\cite{doersch2015unsupervised,dai2016instance,dai2016r}.  Some other
information such as the object boundary can also be explored to guide
the segmentation with more accurate object shape prediction
\cite{chen2016semantic, huang2016object}. 

All the above-mentioned methods strive to improve features and decision
classifiers for better segmentation performance. They
attempt to capture generative object matching templates across training
data. However, their classifiers simply look for the most
{\em likely} patterns with the guidance of the cross-entropy loss in the
softmax-based output layer. This methodology overlooks characteristics
of less common instances, and could be confused by similar patterns of
different classes. In this work, we would like to address this
shortcoming by letting the network learn what does not belong to the
target class as well as better co-existing background/object separation. 

\section{Proposed Reverse Attention Network (RAN)}

\subsection{Motivation}

\begin{figure*}
\begin{center}
\includegraphics[width=12.8cm]{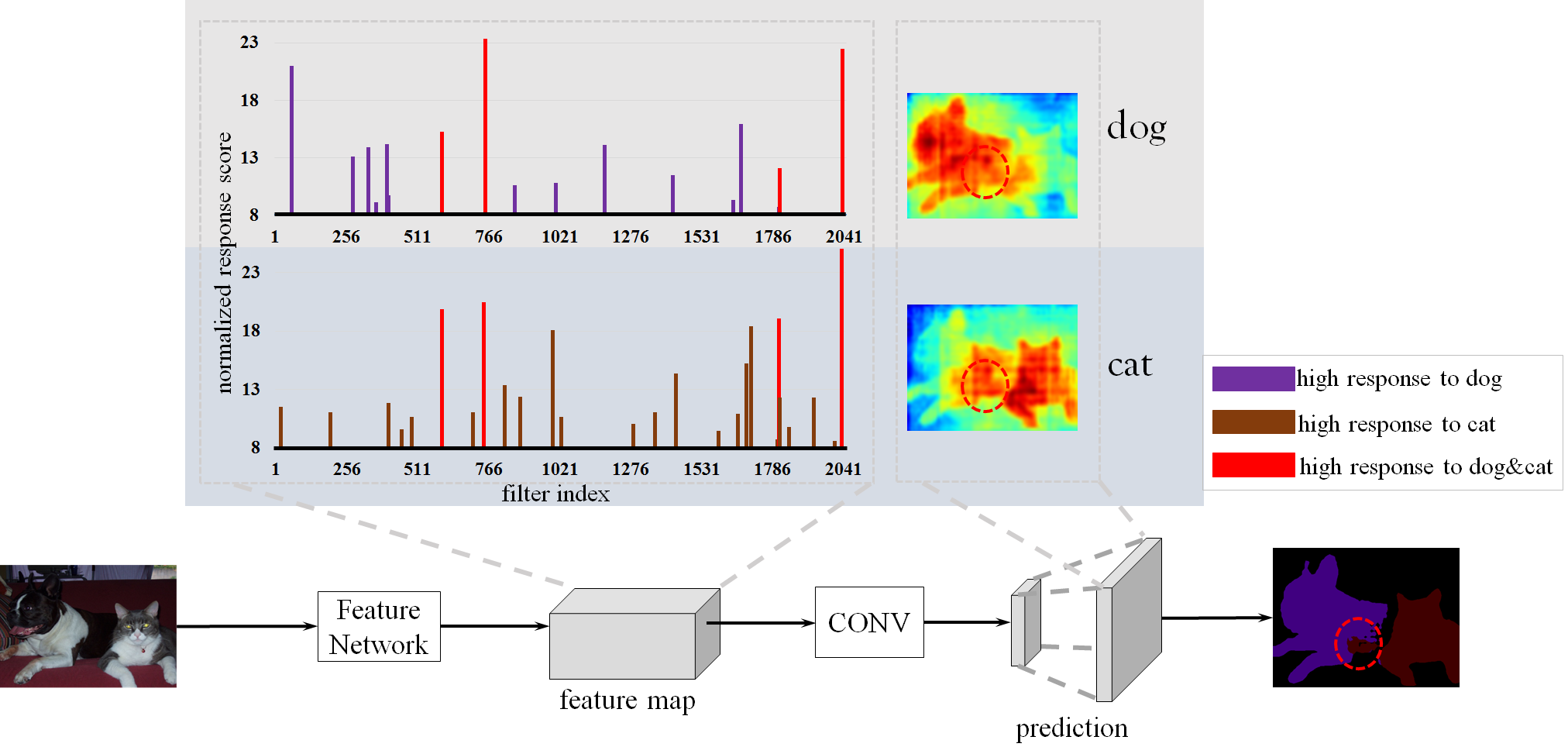}
\end{center}
\caption{Observations on FCN's direct learning. The normalized feature response of the last conv5 layer is presented along with
the class-wise probability map for 'dog' and 'cat'.} \label{fig:observation}
\end{figure*}

Our work is motivated by observations on FCN's learning as given in Fig.
\ref{fig:observation}, where an image is fed into an FCN network. Convolutional
layers of an FCN are usually represented as two parts, the convolutional features network (usually conv1-conv5),
and the class-oriented convolutional layer (CONV) which relates the semantic features to pixel-wise classification results.
Without loss of generality, we use an image that contains a dog and a
cat as illustrated in Fig.  \ref{fig:observation} as an example in our
discussion. 

The segmentation result is shown in the lower-right corner of Fig.
\ref{fig:observation}, where dog's lower body in the circled area is
misclassified as part of a cat.  To explain the phenomenon, we show the
heat maps (i.e. the corresponding filter responses) for the dog and the
cat classes, respectively. It turns out that both classifiers generate
high responses in the circled area.  Classification errors can arise
easily in these confusing areas where two or more classes share similar
spatial patterns. 

To offer additional insights, we plot the normalized filter responses in
the last CONV layer for both classes in Fig.  \ref{fig:observation},
where the normalized response is defined as the sum of all responses of
the same filter per unit area. For ease of visualization, we only show
the filters that have normalized responses higher than a threshold.  The
decision on a target class is primarily contributed by the high response
of a small number of filters while a large number of filters are barely
evoked in the decision. For examples, there are about 20 filters (out of
a total of 2048 filters) that have high responses to the dog or the cat
classes.  We can further divide them into three groups - with a high
response to both the dog and cat classes (in red), with a high response
to the dog class only (in purple) or the cat class (in dark brown) only.
On one hand, these filters, known as the Grand Mother Cell (GMC) filter
\cite{gross2002genealogy,agrawal2014analyzing}, capture the most important semantic patterns of target objects (e.g., the
cat face). On the other hand, some filters have strong responses to
multiple object classes so that they are less useful in discriminating
the underlying object classes. 

Apparently, the FCN is only trained by each class label yet without
being trained to learn the difference between confusing classes.  If we
can let a network learn that the confusing area is not part of a cat
explicitly, it is possible to obtain a network of higher performance.
As a result, this strategy, called the reverse attention learning, may contribute to
better discrimination of confusing classes and better understanding of co-existing
background context in the image. 

\subsection{Proposed RAN System}

\begin{figure*}
\begin{center}
\includegraphics[width=12.8cm]{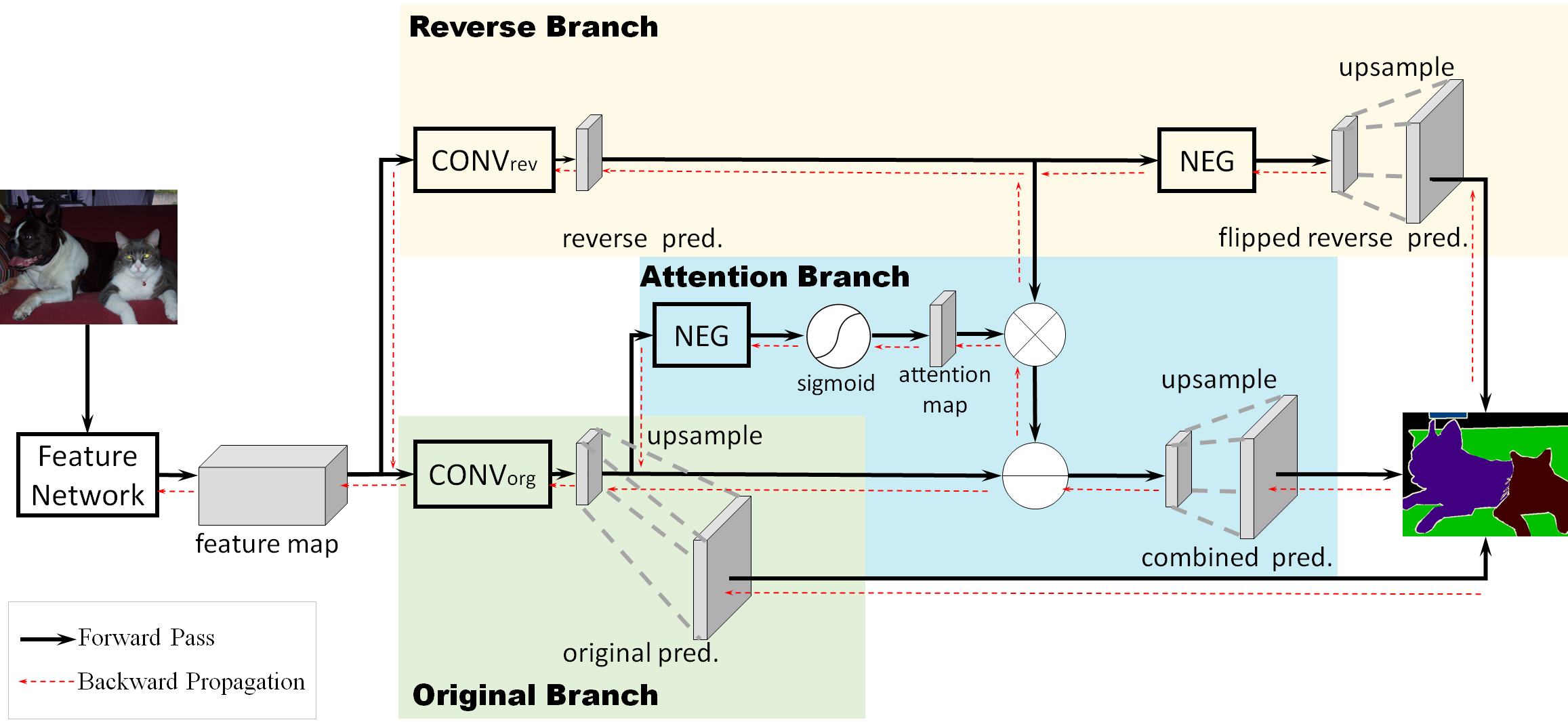}
\end{center}
\caption{The system diagram of the reverse attention network (RAN), where
$CONV_{org}$ and $CONV_{rev}$ filters are used to learn features associated and
not associated with a particular class, respectively.  The reverse object 
class knowledge is then highlighted by an attention mask to generate the
reverse attention of a class, which will then be subtracted from the
original prediction score as a correction.} \label{fig:network}
\end{figure*}

To improve the performance of the FCN, we propose a Reverse Attention
Network (RAN) whose system diagram is depicted in Fig.
\ref{fig:network}.  After getting the feature map, the RAN consists of
three branches: the original branch (the lower path), the attention
branch (the middle path) and the reverse branch (the upper path). The
reverse branch and the attention branch merge to form the reverse
attention response. Finally, decisions from the reverse attention response is subtracted from the
the prediction of original branch
to derive the final decision scores in semantic segmentation. 

The FCN system diagram shown in Fig. \ref{fig:observation} corresponds
to the lower branch in Fig. \ref{fig:network} with the ``original
branch" label. As described earlier, its CONV layers before the feature
map are used to learn object features and its $CONV_{org}$ layers are used
to help decision classifiers to generate the class-wise probability map.
Here, we use $CONV_{org}$ layers to denote that obtained from the original
FCN through a straightforward direct learning process.  For the RAN
system, we introduce two more branches - the reverse branch and the
attention branch. The need of these two branches will be explained
below.

{\bf Reverse Branch.} The upper one in Fig. \ref{fig:network} is the
Reverse Branch. We train another $CONV_{rev}$ layer to learn the reverse
object class explicitly, where the reverse object class is the reversed
ground truth for the object class of concern. In order to obtain the
reversed ground truth, we can set the corresponding class region to zero
and that of the remaining region to one, as illustrated in Fig. \ref{fig:overview}. The remaining region includes
background as well as other classes. However, this would result in specific reverse label for each object class.

There is an alternative way to
implement the same idea. That is, we reverse the sign of all class-wise
response values before feeding them into the softmax-based classifiers.
This operation is indicated by the NEG block in the Reverse Branch. Such
an implementation allows the $CONV_{rev}$ layer to be trained using the same
and original class-wise ground-truth label. 

{\bf Reverse Attention Branch.} 
One simple way to combine results of the original and the reverse
branch is to directly subtract the reverse prediction from the original prediction (in terms of object class
probabilities). We can interpret this operation as finding the
difference between the predicted decision of the original FCN and the
predicted decision due to reverse learning. For example, the lower part
of the dog gives strong responses to both the dog and the cat in the
original FCN. However, the same region will give a strong negative
response to the cat class but almost zero response to the dog class in
the reverse learning branch. Then, the combination of these two branches
will reduce the response to the cat class while preserving the response
to the dog class.

However, directly applying element-wise subtraction does not necessarily result in better performances.
Sometimes the reverse prediction may not do as well as the original prediction in the confident area.
Therefore we propose a reverse attention structure to further highlight the regions which are originally
overlooked in the original prediction, including confusion and background areas. The output of reverse attention structure generates a 
class-oriented mask to amplify the reverse response map. 

As shown in Fig. \ref{fig:network}, the input to the reverse attention branch is the
prediction result of $CONV_{org}$. We flip the sign of the pixel value by
the NEG block, feed the result to the sigmoid function and, finally,
filter the sigmoid output with an attention mask. The sigmoid function
is used to convert the response attention map to the range of [0,1].
Mathematically, the pixel value in the reverse attention map $I_{ra}$ can be
written as
\begin{equation}\label{equ1}
I_{ra}(i,j)= {\rm Sigmoid} (-F_{CONV_{org}}(i,j)),
\end{equation}
where $(i, j)$ denotes the pixel location, and $F_{CONV_{org}}$ 
denotes the response map of $CONV_{org}$, respectively. Note that the region with small or negative responses
$F_{CONV_{org}}$ will be highlighted due to the cascade of the NEG and
the sigmoid operations. In contrast, areas of positive response (or
confident scores) will be suppressed in the reverse attention branch.


After getting the reverse attention map, we combine it with the
$CONV_{rev}$ response map using the element-wise multiplication as shown in
Fig. \ref{fig:network}. The multiplied response score is then subtracted from the original prediction, contributing to our final combined prediction.

Several variants of the RAN architecture have been experimented. The
following normalization strategy offers a faster convergence rate while
providing similar segmentation performance:
\begin{equation}\label{equ2}
I_{ra}(i,j)= {\rm Sigmoid}
(\frac{1}{Relu(F_{CONV_{org}}(i,j))+0.125}-4),
\end{equation}
where $F_{CONV_{org}}$ is normalized to be within $[-4,4]$, which
results in a more uniformed distribution before being fed into the
sigmoid function. Also, we clip all negative scores of $F_{CONV_{org}}$
to zero by applying the Relu operation and control inverse scores to be
within the range of [-4, 4] using parameters $0.125$ and $-4$.  In the
experiment section, we will compare results of the reverse attention
set-ups given in Equations (\ref{equ1}) and (\ref{equ2}). They are 
denoted by RAN-simple (RAN-s) and RAN-normalized (RAN-n), respectively.

{\bf RAN Training.} In order to train the proposed RAN, we
back-propagate the cross-entropy losses at the three branches
simultaneously and adopt the softmax classifiers at the three prediction
outputs.  All three losses are needed to ensure a balanced end-to-end
learning process. The original prediction loss and the reverse
prediction loss allow $CONV_{org}$ and $CONV_{rev}$ to learn the target classes
and their reverse classes in parallel. Furthermore, the loss of the
combined prediction allows the network to learn the reverse attention.
The proposed RAN can be effectively trained based on the pre-trained
FCN, which indicates that the RAN is a further improvement of the FCN by
adding more relevant guidance in the training process.

\section{Experiments}
\begin{table*}[t]
\begin{center}
\resizebox{0.8\textwidth}{!}{%
\begin{tabular}{|c|c|c|c|c|}
\hline
Methods & feature & pixel acc. & mean acc. & mean IoU. \\ \hline \hline
FCN-8s \cite{long2015fully} & \multirow{3}{*}{VGG16} & 65.9 & 46.5 & 35.1 \\
BoxSup \cite{dai2015boxsup} &  & - & - & 40.5 \\
Context \cite{lin2016efficient} &  & 71.5 & 53.9 & 43.3 \\ \hline
VeryDeep \cite{wu2016bridging} & \multirow{3}{*}{ResNet-101} & 72.9 & 54.8 & 44.5 \\
DeepLabv2-ASPP \cite{chen2016deeplab} &  & - & - & 45.7 \\
RefineNet-101 \cite{lin2016refinenet} &  & - & - & 47.1 \\ \hline
Holistic \cite{hu2016recalling} & \multirow{2}{*}{ResNet-152} & 73.5 & 56.6 & 45.8  \\
RefineNet-152 \cite{lin2016refinenet} &  & - & - & 47.3 \\ \hline
Model A2, 2conv \cite{wu2016wider} & Wider ResNet & 75.0 & 58.1 & 48.1 \\ \hline
\begin{tabular}[c]{@{}c@{}}DeepLabv2-LFOV (baseline) \cite{chen2016deeplab}\end{tabular} 
& \multirow{3}{*}{ResNet-101} & - & - & 43.5 \\
RAN-s (ours) &  & 75.3 & 57.1 & \textbf{48.0} \\
RAN-n (ours) &  &  75.3 & 57.2 & \textbf{48.1} \\ \hline
\end{tabular}}
\end{center}
\caption{Comparison of semantic image segmentation performance scores ($\%$) 
on the 5,105 test images of the PASCAL Context dataset.}\label{tab:context}
\end{table*}

To show the effectiveness of the proposed RAN, we conduct experiments on
five datasets. They are the PASCAL Context \cite{mottaghi2014role},
PASCAL Person-Part \cite{chen2014detect}, PASCAL VOC
\cite{Everingham10}, NYU-Depth-v2 \cite{Silberman:ECCV12} and MIT ADE20K
\cite{zhou2016semantic}.  We implemented the RAN using the Caffe
\cite{jia2014caffe} library and built it upon the available DeepLab-v2
repository \cite{chen2016deeplab}. We adopted the initial network
weights provided by the repository, which were pre-trained on 
the COCO dataset with the ResNet-101. All the proposed reverse attention architecture are implemented with the standard Caffe Layers,
where we utilize the $Power Layer$ to flip, shift and scale the response, and use the provided $Sigmoid$
Layer to conduct sigmoid transformation.
 
We employ the "poly" learning rate policy with
$power=0.9$, and basic learning rate equals $0.00025$. Momentum and weight decay are set to 0.9 and 
0.0001 respectively. We adopted the DeepLab data augmentation scheme with random
scaling factor of 0.5, 0.75, 1.0, 1.25, 1.5 and with mirroring for each
training image. Following \cite{chen2016deeplab} we adopt the multi-scale (MSC) input with max fusion
in both training and testing. Although we did not apply the
atrous spatial pyramid pooling (ASPP) due to limited GPU memory, we do
observe significant improvement in the mean intersection-over-union
(mean IoU) score over the baseline DeepLab-v2 LargeFOV and the ASPP
set-up. 

\begin{table*}[t]
\centering
\resizebox{0.75\textwidth}{!}{%
\begin{tabular}{|c|c|c|c|c|c|} 
\hline
Methods & Dil=0 & LargeFOV & +Aug & +MSC & +CRF \\ \hline \hline
DeepLabv2 (baseline) \cite{chen2016deeplab} & 41.6 & 42.6 & 43.2 & 43.5 & 44.4 \\
Dual-Branch RAN     & 42.8 & 43.9 & 44.4 & 45.2 & 46.0 \\
RAN-s              & 44.4 & 45.6 & 46.2 & 47.2 & 48.0 \\
RAN-n               & 44.5 & 45.6 & 46.3 & 47.3 & 48.1 \\ \hline
\end{tabular}}
\caption{Ablation study of different RANs on the PASCAL-Context
dataset to evaluate the benefit of proposed RAN. We compare the results under different network set-up with employing dilated decision conv filters, 
data augmentation, the MSC design and the CRF post-processing.}
\label{tab:ablation}
\end{table*}
{\bf PASCAL-Context.} We first present results conducted on the
challenging PASCAL-Context dataset \cite{mottaghi2014role}. The dataset
has 4,995 training images and 5,105 test images. There are 59 labeled
categories including foreground objects and background context scenes.
We compare the proposed RAN method with a group of state-of-the-art
methods in Table \ref{tab:context}, where RAN-s and RAN-n use
equations (\ref{equ1}) and (\ref{equ2}) in the reverse attention branch,
respectively.  The mean IoU values of RAN-s and RAN-n have a significant improvement
over that of their baseline Deeplabv2-LargeFOV. Our
RAN-s and RAN-n achieve the state-of-the-art mean IoU scores (i.e.,
around 48.1\%) that are comparable with those of the RefineNet
\cite{lin2016refinenet} and the Wider ResNet \cite{wu2016wider}.

\begin{figure}[t]
\centering
\begin{tabular}{cccc}
\subfloat[Image]{\includegraphics[width=2.8cm]{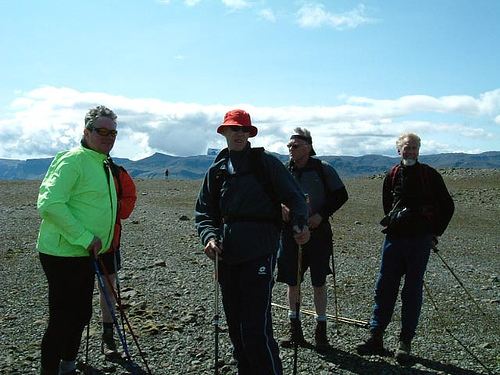}}&
\subfloat[Baseline]{\includegraphics[width=2.8cm]{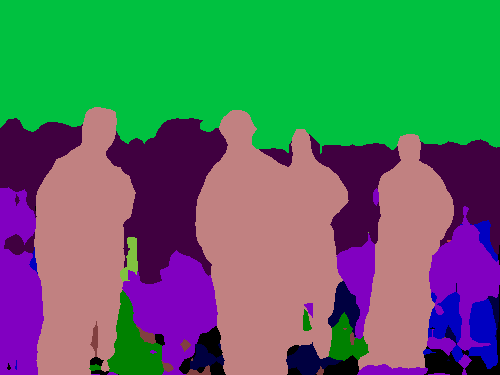}}&
\subfloat[Ours]{\includegraphics[width=2.8cm]{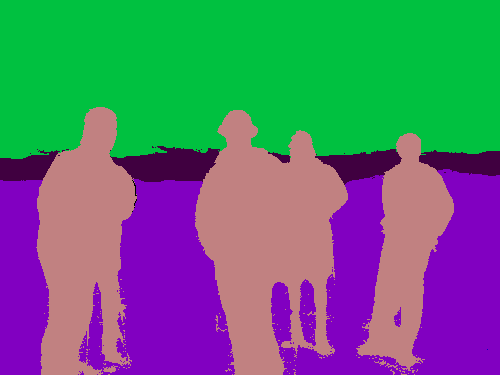}}&
\subfloat[Ground Truth]{\includegraphics[width=2.8cm]{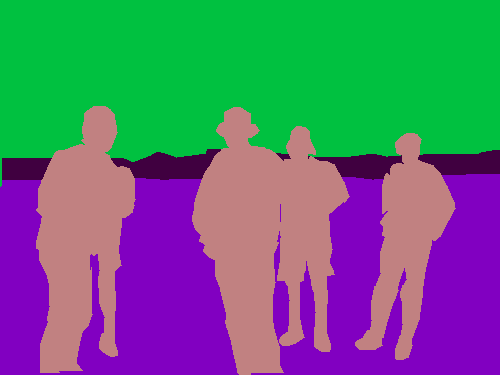}}\\
\subfloat{\includegraphics[width=2.8cm]{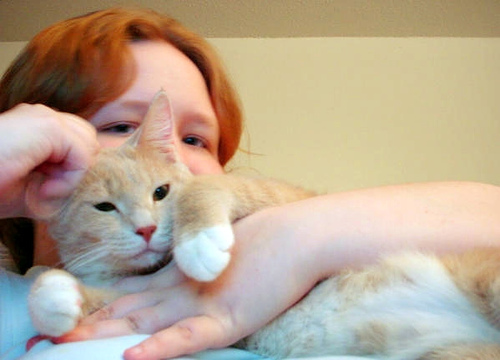}}&
\subfloat{\includegraphics[width=2.8cm]{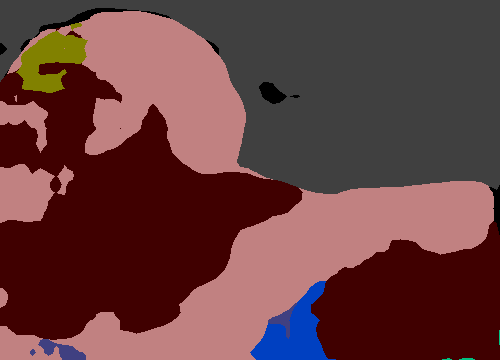}}&
\subfloat{\includegraphics[width=2.8cm]{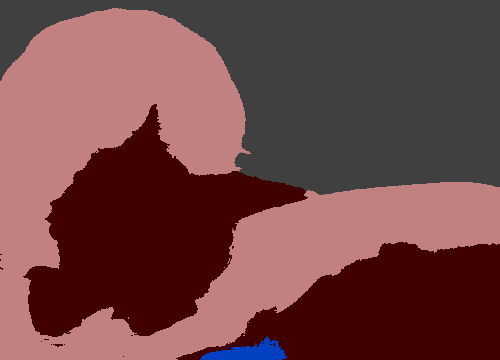}}&
\subfloat{\includegraphics[width=2.8cm]{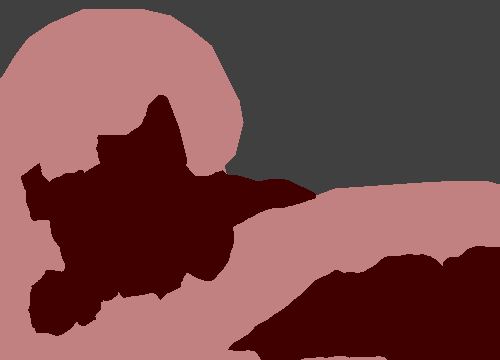}}\\
Image&Baseline&Ours&Ground Truth
\end{tabular}
\caption{Qualitative results in the
PASCAL-Context validation set with: the input image, the DeepLabv2-LargeFOV baseline, our
RAN-s result, and the ground truth.}\label{fig:context}
\end{figure}

We compare the performance of dual-branch RAN (without reverse attention), RAN-s, RAN-n and their baseline DeepLabv2 by conducting a
set of ablation study in Table \ref{tab:ablation}, where a sequence of
techniques is employed step by step. They include dilated classification,
data augmentation, MSC with max fusion and the fully connected conditional random field
(CRF). We see that the performance of RANs keeps improving and they
always outperform their baseline under all situations.  The quantitative
results are provided in Fig.  \ref{fig:context}. It shows that the
proposed reverse learning can correct some mistakes in the confusion
area, and results in more uniformed prediction for the target object. 
 
{\bf PASCAL Person-Part.} We also conducted experiments on the
PASCAL Person-Part dataset \cite{chen2014detect}. It includes labels of
six body parts of persons (i.e., Head, Torso, Upper/Lower Arms and
Upper/Lower Legs).  There are 1,716 training images and 1,817 validation
images. As observed in \cite{chen2016deeplab}, the dilated decision
classifier provides little performance improvement.  Thus, we also adopted
the MSC structure with 3-by-3 decision filters without dialtion for RANs. The mean IoU
results of several benchmarking methods are shown in Table
\ref{tab:person}.The results demonstrate that both RAN-s and RAN-n outperform the baseline
DeepLabv2 and achieves state-of-the-art performance in this fine-grained dataset. 

\begin{table}[h]
\centering
\resizebox{0.9\textwidth}{!}{%
\begin{tabular}{|c|c|c|c|c|c|c|c|}
\hline
 & Attention \cite{chen2016attention} & HAZN \cite{xia2016zoom} & Graph LSTM \cite{liang2016semantic} & 
RefineNet \cite{lin2016refinenet} & DeepLabv2 \cite{chen2016deeplab} & RAN-s & RAN-n \\ \hline \hline
mean IoU & 56.4 & 57.5 & 60.2 & 68.6 & 64.9 & 66.6 & 66.5 \\ \hline
\end{tabular}}
\caption{Comparison of the mean IoU scores ($\%$) of several benchmarking 
methods for the PASCAL PERSON-Part dataset.}\label{tab:person}
\end{table}

{\bf PASCAL VOC2012.} Furthermore, we conducted experiments on the popular
PASCAL VOC2012 test set \cite{Everingham10}. We adopted the augmented
ground truth from \cite{hariharan2014simultaneous} with a total of
12,051 training images and submitted our segmentation results to the
evaluation website. We find that for the VOC dataset, our DeepLab based network
does not have significant improvement as the specifically designed networks such as 
\cite{lin2016refinenet,zhao2016pyramid}.  However we still observer
about $1.4\%$ improvement over the baseline DeepLabv2-LargeFOV,
which also outperforms the DeepLabv2-ASPP. 

\begin{table}[h]
\centering
\resizebox{\textwidth}{!}{%
\begin{tabular}{|c|cccccccccccccccccccc||c|}
\hline
Method & aero & bike & bird & boat & bottle & bus & car & cat & chair & cow & table & dog & horse & mbike & person & potted & sheep & sofa & train & tv & mean \\  \hline \hline
FCN-8s \cite{long2015fully} & 76.8 & 34.2 & 68.9 & 49.4 & 60.3 & 75.3 & 74.7 & 77.6 & 21.4 & 62.5 & 46.8 & 71.8 & 63.9 & 76.5 & 73.9 & 45.2 & 72.4 & 37.4 & 70.9 & 55.1 & 62.2 \\ \hline
Context \cite{lin2016efficient} & 94.1 & 40.7 & 84.1 & 67.8 & 75.9 & 93.4 & 84.3 & 88.4 & 42.5 & 86.4 & 64.7 & 85.4 & 89.0 & 85.8 & 86.0 & 67.5 & 90.2 & 63.8 & 80.9 & 73.0 & 78.0 \\ \hline
VeryDeep \cite{wu2016bridging} & 91.9 & 48.1 & 93.4 & 69.3 & 75.5 & 94.2 & 87.5 & 92.8 & 36.7 & 86.9 & 65.2 & 89.1 & 90.2 & 86.5 & 87.2 & 64.6 & 90.1 & 59.7 & 85.5 & 72.7 & 79.1 \\ \hline
DeepLabv2-LFOV \cite{chen2016deeplab} & 93.0 & 41.6 & 91.0 & 65.3 & 74.5 & 94.2 & 88.8 & 91.7 & 37.2 & 87.9 & 64.6 & 89.7 & 91.8 & 86.7 & 85.8 & 62.6 & 88.6 & 60.1 & 86.6 & 75.4 & 79.1 \\ \hline
DeepLabv2-ASPP \cite{chen2016deeplab} & 92.6 & 60.4 & 91.6 & 63.4 & 76.3 & 95.0 & 88.4 & 92.6 & 32.7 & 88.5 & 67.6 & 89.6 & 92.1 & 87.0 & 87.4 & 63.3 & 88.3 & 60.0 & 86.8 & 74.5 & 79.7 \\ \hline
RAN-s$^{1}$ & 92.7 & 44.7 & 91.9 & 68.2 & 79.3 & 95.4 & 91.2 & 93.3 & 42.8 & 87.8 & 66.9 & 89.1 & 93.2 & 89.5 & 88.4 & 61.6 & 89.8 & 62.6 & 87.8 & 77.8 & 80.5 \\ \hline
RAN-n$^{2}$ & 92.5 & 44.6 & 92.1 & 68.8 & 79.1 & 95.5 & 91.0 & 93.1 & 43.1 & 88.3 & 66.6 & 88.9 & 93.4 & 89.3 & 88.3 & 61.2 & 89.7 & 62.5 & 87.7 & 77.6 & 80.4 \\ \hline
\end{tabular}}
{\tiny$^{1}$\url{http://host.robots.ox.ac.uk:8080/anonymous/QHUF8T.html},
$^{2}$\url{http://host.robots.ox.ac.uk:8080/anonymous/UWJO3S.html}}
\caption{Comparison of the mean IoU scores ($\%$) per object class of
several methods for the PASCAL VOC2012 test dataset.}\label{tab:voc}
\end{table}
\begin{figure*}[t]
\centering
\begin{tabular}{cccc}
\subfloat[Image]{\includegraphics[width=2.8cm]{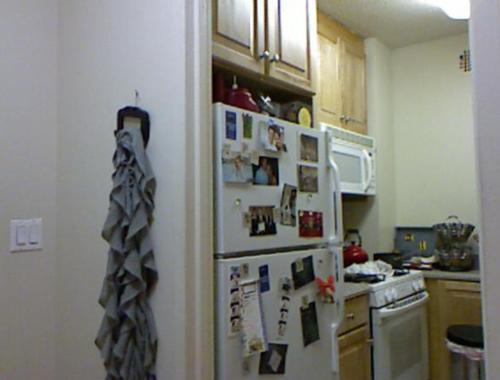}}&
\subfloat[Baseline]{\includegraphics[width=2.8cm]{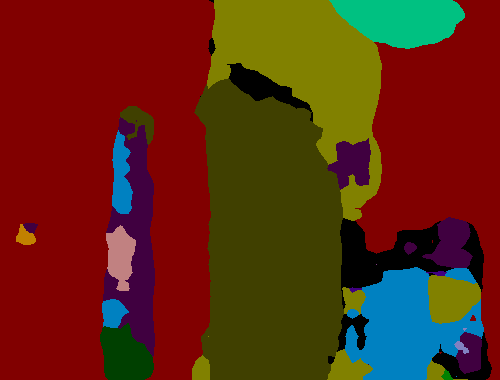}}&
\subfloat[Ous]{\includegraphics[width=2.8cm]{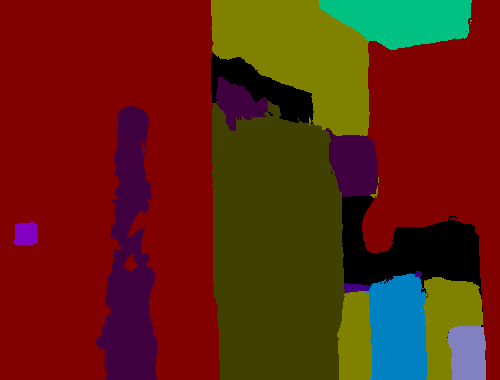}}&
\subfloat[Ground Truth]{\includegraphics[width=2.8cm]{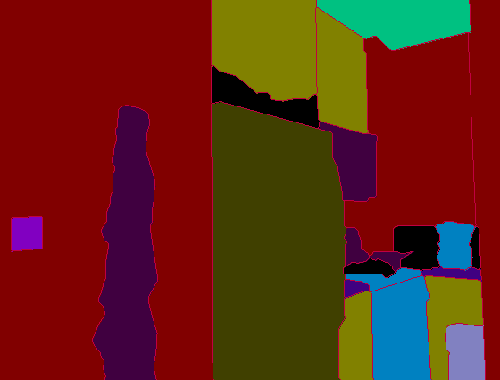}}\\
\subfloat{\includegraphics[width=2.8cm]{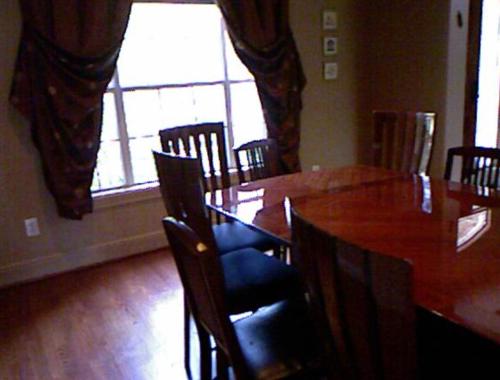}}&
\subfloat{\includegraphics[width=2.8cm]{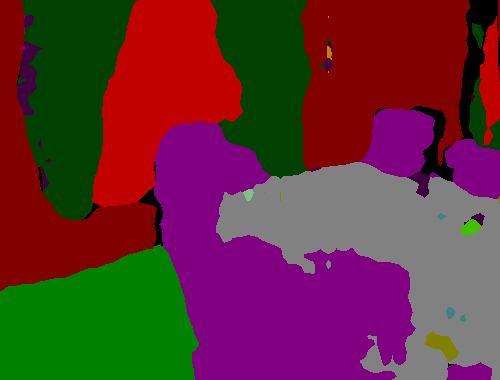}}&
\subfloat{\includegraphics[width=2.8cm]{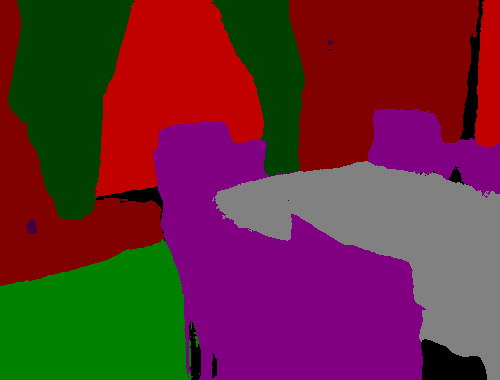}}&
\subfloat{\includegraphics[width=2.8cm]{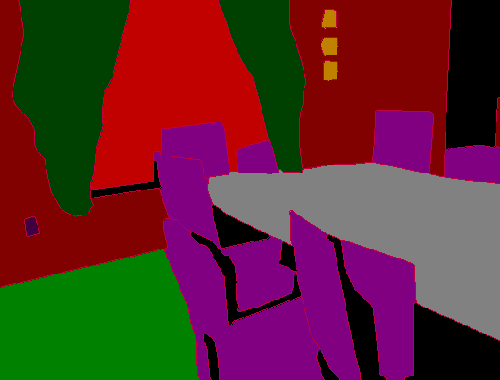}}\\
Image&Baseline&Ours&Ground Truth
\end{tabular}
\caption{Qualitative results in the
NYU-DepthV2 validation set with: the input image, the DeepLabv2-LargeFOV baseline, our
RAN-s result, and the ground truth.}\label{fig:nyu}
\end{figure*}

{\bf NYUDv2.} The NYUDv2 dataset \cite{Silberman:ECCV12} is an indoor
scene dataset with 795 training images and 654 test images. It has
coalesced labels of 40 classes provided by \cite{gupta2013perceptual}.
The mean IoU results of several benchmarking methods are shown in Table
\ref{tab:nyu}.  We see that RAN-s and RAN-n improve their baseline
DeepLabv2-LargeFOV by a large margin (around 3\%). Visual comparison
of segmentation results of two images are shown in Fig. \ref{fig:nyu}.

\begin{table}[h]
\centering
\resizebox{0.98\textwidth}{!}{%
\begin{tabular}{|c|c|c|c|c|c|c|c|c|c|}
\hline
 & Gupta et al. \cite{gupta2014learning} & FCN-32s \cite{long2015fully} 
& Context \cite{lin2016efficient} & Holistic \cite{hu2016recalling} 
& RefineNet \cite{lin2016refinenet} & DeepLabv2-ASPP \cite{chen2016deeplab} 
& DeepLabv2-LFOV \cite{chen2016deeplab} & RAN-s & RAN-n \\  \hline \hline
feature & \multicolumn{3}{c|}{VGG16} & \multicolumn{2}{c|}{ResNet-152} & \multicolumn{4}{c|}{ResNet-101} \\ \hline
mean IoU & 28.6 & 29.2 & 40.6 & 38.8 & 46.5 & 37.8 & 37.3 & 41.2 & 40.7 \\ \hline
\end{tabular}}
\caption{Comparison of the mean IoU scores ($\%$) of several benchmarking
methods on the NYU-Depth2 dataset.}\label{tab:nyu}
\end{table}


{\bf MIT ADE20K.} The MIT ADE20K dataset \cite{zhou2016semantic} was
released recently. The dataset has 150 labeled classes for both objects
and background scene parsing. There are about 20K and 2K images in the
training and validation sets, respectively. Although our baseline DeepLabv2 does not perform well in global
scene parsing as in 
\cite{hu2016recalling, zhao2016pyramid}, we still observe about 2\% improvement in the
mean IoU score as shown in Table \ref{tab:ade}. 

\begin{table}[h]
\centering
\resizebox{0.98\textwidth}{!}{%
\begin{tabular}{|c|c|c|c|c|c|c|c|c|c|}
\hline
 & FCN-8s \cite{zhou2016semantic} & DilatedNet \cite{zhou2016semantic} 
& DilatedNet Cascade \cite{zhou2016semantic} & Holistic \cite{hu2016recalling}
& PSPNet \cite{zhao2016pyramid} & DeepLabv2-ASPP \cite{chen2016deeplab} 
& DeepLabv2-LFOV \cite{chen2016deeplab} & RAN-s & RAN-n \\  \hline \hline
feature & \multicolumn{1}{c|}{VGG16} & \multicolumn{2}{c|}{ResNet-101} & \multicolumn{2}{c|}{ResNet-152} & \multicolumn{4}{c|}{ResNet-101} \\ \hline
mean IoU & 29.39 & 32.31 & 34.9 & 37.93 & 43.51 & 34.0 & 33.1 & 35.2 & 35.3 \\ \hline
\end{tabular}}
\caption{Comparison of the mean IoU scores ($\%$) of several benchmarking
methods on the ADE20K dataset.}\label{tab:ade}
\end{table}

\section{Conclusion}

A new network, called the RAN, designed for reverse learning was
proposed in this work.  The network explicitly learns what are and are
not associated with a target class in its direct and reverse branches,
respectively.  To further enhance the reverse learning effect, the
sigmoid activation function and an attention mask were introduced to
build the reverse attention branch as the third one.  The three branches
were integrated in the RAN to generate final results.  The RAN provides
significant performance improvement over its baseline network and
achieves the state-of-the-art semantic segmentation performance in
several benchmark datasets.

\bibliographystyle{splncs}
\bibliography{egbib}

\begin{thebibliography}{10}

\bibitem{long2015fully}
Long, J., Shelhamer, E., Darrell, T.:
\newblock Fully convolutional networks for semantic segmentation.
\newblock In: Proceedings of the IEEE Conference on Computer Vision and Pattern
  Recognition. (2015)  3431--3440

\bibitem{simonyan2014very}
Simonyan, K., Zisserman, A.:
\newblock Very deep convolutional networks for large-scale image recognition.
\newblock arXiv preprint arXiv:1409.1556 (2014)

\bibitem{szegedy2015going}
Szegedy, C., Liu, W., Jia, Y., Sermanet, P., Reed, S., Anguelov, D., Erhan, D.,
  Vanhoucke, V., Rabinovich, A.:
\newblock Going deeper with convolutions.
\newblock In: Proceedings of the IEEE Conference on Computer Vision and Pattern
  Recognition. (2015)  1--9

\bibitem{he2015deep}
He, K., Zhang, X., Ren, S., Sun, J.:
\newblock Deep residual learning for image recognition.
\newblock arXiv preprint arXiv:1512.03385 (2015)

\bibitem{bishop2006pattern}
Bishop, C.M.:
\newblock Pattern recognition and machine learning.
\newblock springer (2006)

\bibitem{lin2016refinenet}
Lin, G., Milan, A., Shen, C., Reid, I.:
\newblock Refinenet: Multi-path refinement networks with identity mappings for
  high-resolution semantic segmentation.
\newblock arXiv preprint arXiv:1611.06612 (2016)

\bibitem{zhao2016pyramid}
Zhao, H., Shi, J., Qi, X., Wang, X., Jia, J.:
\newblock Pyramid scene parsing network.
\newblock arXiv preprint arXiv:1612.01105 (2016)

\bibitem{hu2016recalling}
Hu, H., Deng, Z., Zhou, G.T., Sha, F., Mori, G.:
\newblock Recalling holistic information for semantic segmentation.
\newblock arXiv preprint arXiv:1611.08061 (2016)

\bibitem{doersch2015unsupervised}
Doersch, C., Gupta, A., Efros, A.A.:
\newblock Unsupervised visual representation learning by context prediction.
\newblock In: Proceedings of the IEEE International Conference on Computer
  Vision. (2015)  1422--1430

\bibitem{jia2014caffe}
Jia, Y., Shelhamer, E., Donahue, J., Karayev, S., Long, J., Girshick, R.,
  Guadarrama, S., Darrell, T.:
\newblock Caffe: Convolutional architecture for fast feature embedding.
\newblock In: Proceedings of the ACM International Conference on Multimedia,
  ACM (2014)  675--678

\bibitem{zhang1996survey}
Zhang, Y.J.:
\newblock A survey on evaluation methods for image segmentation.
\newblock Pattern recognition \textbf{29} (1996)  1335--1346

\bibitem{shi2000normalized}
Shi, J., Malik, J.:
\newblock Normalized cuts and image segmentation.
\newblock Pattern Analysis and Machine Intelligence, IEEE Transactions on
  \textbf{22} (2000)  888--905

\bibitem{krizhevsky2012imagenet}
Krizhevsky, A., Sutskever, I., Hinton, G.E.:
\newblock Imagenet classification with deep convolutional neural networks.
\newblock In: Advances in neural information processing systems. (2012)
  1097--1105

\bibitem{Everingham10}
Everingham, M., Van~Gool, L., Williams, C.K.I., Winn, J., Zisserman, A.:
\newblock The pascal visual object classes (voc) challenge.
\newblock International Journal of Computer Vision \textbf{88} (2010)  303--338

\bibitem{chen2016attention}
Chen, L.C., Yang, Y., Wang, J., Xu, W., Yuille, A.L.:
\newblock Attention to scale: Scale-aware semantic image segmentation.
\newblock In: Proceedings of the IEEE Conference on Computer Vision and Pattern
  Recognition. (2016)  3640--3649

\bibitem{chen2016deeplab}
Chen, L.C., Papandreou, G., Kokkinos, I., Murphy, K., Yuille, A.L.:
\newblock Deeplab: Semantic image segmentation with deep convolutional nets,
  atrous convolution, and fully connected crfs.
\newblock arXiv preprint arXiv:1606.00915 (2016)

\bibitem{zheng2015conditional}
Zheng, S., Jayasumana, S., Romera-Paredes, B., Vineet, V., Su, Z., Du, D.,
  Huang, C., Torr, P.H.:
\newblock Conditional random fields as recurrent neural networks.
\newblock In: Proceedings of the IEEE International Conference on Computer
  Vision. (2015)  1529--1537

\bibitem{shrivastava2016training}
Shrivastava, A., Gupta, A., Girshick, R.:
\newblock Training region-based object detectors with online hard example
  mining.
\newblock arXiv preprint arXiv:1604.03540 (2016)

\bibitem{dai2016instance}
Dai, J., He, K., Li, Y., Ren, S., Sun, J.:
\newblock Instance-sensitive fully convolutional networks.
\newblock arXiv preprint arXiv:1603.08678 (2016)

\bibitem{dai2016r}
Dai, J., Li, Y., He, K., Sun, J.:
\newblock R-fcn: Object detection via region-based fully convolutional
  networks.
\newblock arXiv preprint arXiv:1605.06409 (2016)

\bibitem{chen2016semantic}
Chen, L.C., Barron, J.T., Papandreou, G., Murphy, K., Yuille, A.L.:
\newblock Semantic image segmentation with task-specific edge detection using
  cnns and a discriminatively trained domain transform.
\newblock In: Proceedings of the IEEE Conference on Computer Vision and Pattern
  Recognition. (2016)  4545--4554

\bibitem{gross2002genealogy}
Gross, C.G.:
\newblock Genealogy of the “grandmother cell”.
\newblock The Neuroscientist \textbf{8} (2002)  512--518

\bibitem{agrawal2014analyzing}
Agrawal, P., Girshick, R., Malik, J.:
\newblock Analyzing the performance of multilayer neural networks for object
  recognition.
\newblock In: European Conference on Computer Vision, Springer (2014)  329--344

\bibitem{dai2015boxsup}
Dai, J., He, K., Sun, J.:
\newblock Boxsup: Exploiting bounding boxes to supervise convolutional networks
  for semantic segmentation.
\newblock In: Proceedings of the IEEE International Conference on Computer
  Vision. (2015)  1635--1643

\bibitem{lin2016efficient}
Lin, G., Shen, C., van~den Hengel, A., Reid, I.:
\newblock Efficient piecewise training of deep structured models for semantic
  segmentation.
\newblock In: Proceedings of the IEEE Conference on Computer Vision and Pattern
  Recognition. (2016)  3194--3203

\bibitem{wu2016bridging}
Wu, Z., Shen, C., Hengel, A.v.d.:
\newblock Bridging category-level and instance-level semantic image
  segmentation.
\newblock arXiv preprint arXiv:1605.06885 (2016)

\bibitem{wu2016wider}
Wu, Z., Shen, C., Hengel, A.v.d.:
\newblock Wider or deeper: Revisiting the resnet model for visual recognition.
\newblock arXiv preprint arXiv:1611.10080 (2016)

\bibitem{mottaghi2014role}
Mottaghi, R., Chen, X., Liu, X., Cho, N.G., Lee, S.W., Fidler, S., Urtasun, R.,
  Yuille, A.:
\newblock The role of context for object detection and semantic segmentation in
  the wild.
\newblock In: Proceedings of the IEEE Conference on Computer Vision and Pattern
  Recognition. (2014)  891--898

\bibitem{chen2014detect}
Chen, X., Mottaghi, R., Liu, X., Fidler, S., Urtasun, R., Yuille, A.:
\newblock Detect what you can: Detecting and representing objects using
  holistic models and body parts.
\newblock In: Proceedings of the IEEE Conference on Computer Vision and Pattern
  Recognition. (2014)  1971--1978

\bibitem{Silberman:ECCV12}
Nathan~Silberman, Derek~Hoiem, P.K., Fergus, R.:
\newblock Indoor segmentation and support inference from rgbd images.
\newblock In: ECCV. (2012)

\bibitem{zhou2016semantic}
Zhou, B., Zhao, H., Puig, X., Fidler, S., Barriuso, A., Torralba, A.:
\newblock Semantic understanding of scenes through the ade20k dataset.
\newblock arXiv preprint arXiv:1608.05442 (2016)

\bibitem{xia2016zoom}
Xia, F., Wang, P., Chen, L.C., Yuille, A.L.:
\newblock Zoom better to see clearer: Human and object parsing with
  hierarchical auto-zoom net.
\newblock In: European Conference on Computer Vision, Springer (2016)  648--663

\bibitem{liang2016semantic}
Liang, X., Shen, X., Feng, J., Lin, L., Yan, S.:
\newblock Semantic object parsing with graph lstm.
\newblock In: European Conference on Computer Vision, Springer (2016)  125--143

\bibitem{hariharan2014simultaneous}
Hariharan, B., Arbel{\'a}ez, P., Girshick, R., Malik, J.:
\newblock Simultaneous detection and segmentation.
\newblock In: Computer vision--ECCV 2014.
\newblock Springer (2014)  297--312

\bibitem{gupta2013perceptual}
Gupta, S., Arbelaez, P., Malik, J.:
\newblock Perceptual organization and recognition of indoor scenes from rgb-d
  images.
\newblock In: Proceedings of the IEEE Conference on Computer Vision and Pattern
  Recognition. (2013)  564--571

\bibitem{gupta2014learning}
Gupta, S., Girshick, R., Arbel{\'a}ez, P., Malik, J.:
\newblock Learning rich features from rgb-d images for object detection and
  segmentation.
\newblock In: European Conference on Computer Vision, Springer (2014)  345--360

\end{thebibliography}

\end{document}